\documentclass[10pt,twocolumn,letterpaper]{article}

\usepackage{wacv}
\usepackage{times}
\usepackage{epsfig}
\usepackage{graphicx}
\usepackage{amsmath}
\usepackage{amssymb}

\def\wacvPaperID{763} 

\wacvfinalcopy 
\ifwacvfinal
\fi

\ifwacvfinal
\usepackage[breaklinks=true,bookmarks=false]{hyperref}
\else
\usepackage[pagebackref=true,breaklinks=true,colorlinks,bookmarks=false]{hyperref}
\fi

\ifwacvfinal
\else

\fi

\begin{document}

\title{Line Art Correlation Matching Feature Transfer Network for Automatic Animation Colorization}
\author{Qian Zhang, Bo Wang, Wei Wen, Hai Li, Junhui Liu\\
iQIYI Inc\\
Chengdu, China\\
{\tt\small \{zhangqian07, wangbo03, wenwei, lihai, liujunhui\}@qiyi.com}
}
\maketitle

\begin{abstract}
  Automatic animation line art colorization is a challenging computer vision problem, since the information of the line art is highly sparse and abstracted and there exists a strict requirement for the color and style consistency between frames. Recently, a lot of Generative Adversarial Network (GAN) based image-to-image translation methods for single line art colorization have emerged. They can generate perceptually appealing results conditioned on line art images. However, these methods can not be adopted for the purpose of animation colorization because there is a lack of consideration of the in-between frame consistency.
Existing methods simply input the previous colored frame as a reference to color the next line art, which will mislead the colorization due to the spatial misalignment of the previous colored frame and the next line art especially at positions where apparent changes happen. To address these challenges, we design a kind of correlation matching feature transfer model (called CMFT) to align the colored reference feature in a learnable way and integrate the model into an U-Net based generator in a coarse-to-fine manner. This enables the generator to transfer the layer-wise synchronized features from the deep semantic code to the content progressively. Extension evaluation shows that CMFT model can effectively improve the in-between consistency and the quality of colored frames especially when the motion is intense and diverse.
\end{abstract}

\begin{figure*}
\begin{center}
  \includegraphics[width=\textwidth]{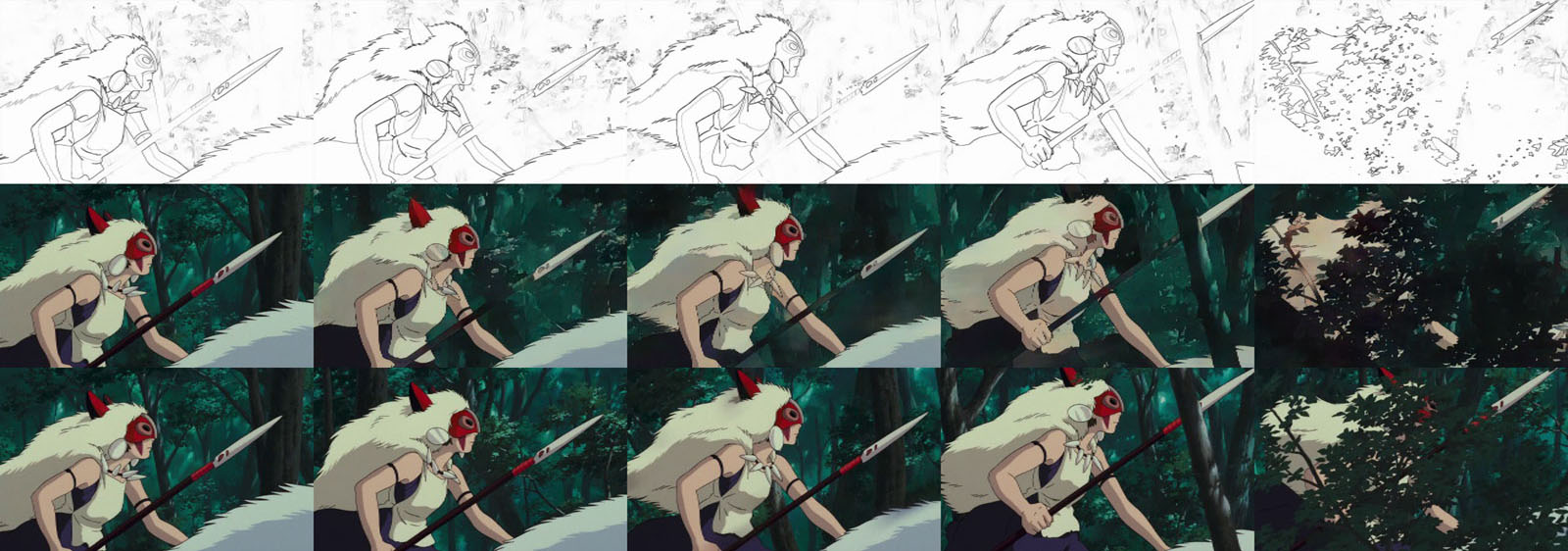}
\end{center}
  \caption{Animation sequence colorization example. The first row shows a sketch sequence, the second row shows the colored frames generated by our network conditioned on the
colored frame in the first column, the third row shows the original animation frames. The example sequence is from the film \textit{Princess Mononoke}.}
  \label{fig:teaser}
\end{figure*}
\section{Introduction}

Nowadays, animation has become part of our daily entertainments, thousands of animations have accounted for a large proportion of the global viewership both on TV and online video platforms.  
According to the AJA's 2018 report \cite{AIR18}, the popularity of animation is still growing. 
Six years of continuous growth has been seen in Japan's anime market. 
However, animation production is a complex and time-consuming process that requires a large number of workers to collaborate in different stages. The key-frame sketches that define the major character movements are portrayed by lead artists while the in-between sketches of motions are completed by inexperienced artists. Then, the labor workers repetitively colorize all the line arts on the basis of the character's color chart previously designed by the lead artists. This colorization procedure is considered to be a tedious and labor-intensive work. Thus, finding an automatic method to consistently colorize the sketch frames can significantly improve the efficiency of animation production and greatly save the expenses and labour cost.


The image-to-image translation method presented by Isola et al \cite{isola2017image} utilize Generative Adversarial Networks (GANs) to learn a mapping model from the source image domain to the target image domain. 
The similar idea has been applied to various tasks such as generating photographs from attribute and semantic distributions. 
There has been a lot of learning based methods \cite{isola2017image,zhu2017unpaired,chen2018sketchygan,paintschainer,liu2017auto,kim2019tag2pix,frans2017outline,zhang2017style,furusawa2017comicolorization,hensman2017cgan,ci2018user,zhang2018two} for single sketch colorization, most of which treat the problem as an image-to-image translation task, aiming at generating a perceptual pleasing result. 
However, due to the lack of consideration of the in-between consistency, this kind of methods can not be directly adopted to colorize frame sequences.

In \cite{thasarathan2019automatic}, temporal informations are incorporated into the image-to-image translation network to encourage the consistency between colorized frames by simply taking the previous colored frame as an input to predict the current colored frame. 
Two problems exist in these methods. 
Firstly, the semantic distribution between the previous colored frame and the current sketch frame is misaligned in the spatial domain, which will mislead the colorization, especially at positions where apparent changes happen. 
Secondly, although information of the previous colored frame and the current sketch frame is used to do prediction, information of the previous sketch which is highly related to both the previous colored frame and the current sketch is ignored. 

To address the above problems, we propose a coherent line art colorization framework with a learnable correlation matching feature transfer model (called CMFT) to match the correlation in feature maps. 
The CMFT model utilizes two kinds of consistencies of four frames, which consist of the domain style consistency and the spatial content consistency. 
On the one hand, because of the domain style consistency between the previous and next line art, the in-between content transformation can be presented by the correlation of continuous frames. 
On the other hand, because of the spatial content consistency of the next line art and colored image, we assume that the in-between motion can be maintained across two style domains obviously. 
Therefore, the transformation can be applied to color domain to reconstruct the target image from semantic or texture patches of the previous color image. 
To simulate the animation colorization behviour that artists usually determine the global color composition before local details, we integrate a series of CMFT models into a coarse-to-fine decoder. 
Simultaneously, we introduct a network to decrease the matching difficulty brought by the serious sparsity of line art.
Overall, our contributions are as follows:
\begin{itemize}
\item We propose a learnable CMFT model to reconstruct the target color image by matching the correlation of feature maps and applying the in-between motion to the color domain.
\item We design a coherent line art sequence colorization framework consisting of four encoders and one decoder, which can generate high-quality colorized images effectively and efficiently.
\item We devise a method to build diverse and discriminative dataset from cartoon films for the coherent frame sequence colorization task. 
\end{itemize}

\section{Related work}
\subsection{Sketch Line Art Colorization}

Recently, GAN\cite{radford2015unsupervised} has offered superior quality in generation tasks compared to conventional image generation methods. 
Several studies have been conducted on GAN for line art colorization, which train CNNs on large datasets to combine low-level local details and high-level semantic information to produce a perpetual appealing image. 
Isola et al\cite{isola2017image}, Zhu et al \cite{zhu2017unpaired} and Chen et al \cite{chen2018sketchygan} learn a direct mapping from human drawn sketches (for a particular category or with category labels) to realistic images with generative adversarial networks. 
PaintChainer \cite{paintschainer} develops an online application that can generate pleasing colorization results for anime line arts based on an U-Net based generator. 
\cite{ci2018user} improves colorization quality by adding an independent local feature network to the generator. 
To increase the color diversity and control the style of image, reference are added to the generator.  
In \cite{ci2018user,zhang2018two}, points or lines with specified colors are input to a generator as hints to change color layouts of the target drawing positions or areas. 
In \cite{furusawa2017comicolorization}, a color palette is used to guide the color distribution of the result. 
In \cite{zhang2017style}, the VGG features of the sample image is added to the generator as a style hint. 
Style2Paints \cite{zhang2018two} extends the method by adding a refinement stage, which provides a state-of-the-art result in single sketch colorization. 

Howerver, none of these works can be directly transplanted to the frame sequence colorization.  
Since no meticulous-designed dense reference has been introduced to affect details of the result, rigid color consistency required in the frame sequence colorization task can not be well guranteed. 
Thasarathan et al \cite{thasarathan2019automatic} is the first study working on colorizing sketch frame sequences, which takes the previous colored image as a dense reference and simply concatenates it with the sketch as an input of the encoder.
This will mislead the colorization because of the spatial misalignment between a sketch and the corresponding color reference. 
In this paper, we reconstruct the aligned color reference by finding the correlation of sketch features.

\subsection{Traditional Sketch Correlation}
Despite a strong correlation between every two adjacent frames, finding the correction between sketches is a difficult task, because features of sketches are sparse and highly-abstracted. 
Some studies \cite{song2013automatic,xing2015autocomplete,zhu2016globally,sato2014reference} assume that the line art is closed and can be segmented into different shape areas, and they use shapes and topological features to find the correlation between adjacent sketches. 
Some other studies \cite{sykora2009rigid,sykora2011textoons,noris2011temporal} are proposed to model the correspondence between two frames as a as-rigid-as-possible deformation, which is interatively found by matching local features. 
Those methods can not handle complex sketch changes, because they depend on the stability of shapes, topology or local feautres, which often varies from adjacent animation frames. \\

\subsection{Deep CNN Feature Matching based Transfer}
Another way to find the correspondence between images is deep feature matching. 
Local patches in deep features have characteristic arrangements of feature activations to describe objects, and higher-up code becomes more invariant under in-class variation \cite{li2016combining}. 
It has been shown in high-level image recognition tasks that such deep features are better representations for images \cite{zeiler2014visualizing}. 
Li et al \cite{li2016combining} realizes the image-to-image translation between photograph and style image via matching local patches of features extracted from a pre-trained VGG network. 
In order to transfer an image $I$ of domain $A$ to domain $B$, the features of domain $B$ is aligned to the content of image $I$ by matching the patches of deep features, and then the transferred image is reconstructed from aligned features. 
Liao et al \cite{liao2017visual} formulates the transfer mapping as a problem of image analogies \cite{hertzmann2001image,cheng2008consistent} by seperating the matching into one in-place mapping (spatial invariant) and one similar-appearance mapping (style invariant) to improve the transfer quality and presicion. 
The pre-trained VGG network can offer adequate semantics for correct patch matching, but it only adapts general photographs instead of sparse and highly-abstracted sketch representations. 
In order to learn effective sketch features, we design a learnable correlation matching model and integrate it to our generator for training. 
This module will guide the network to learn a good representations for the sketch frame sequence colorization task by itself.
\begin{figure}[h]
\includegraphics[width=\linewidth]{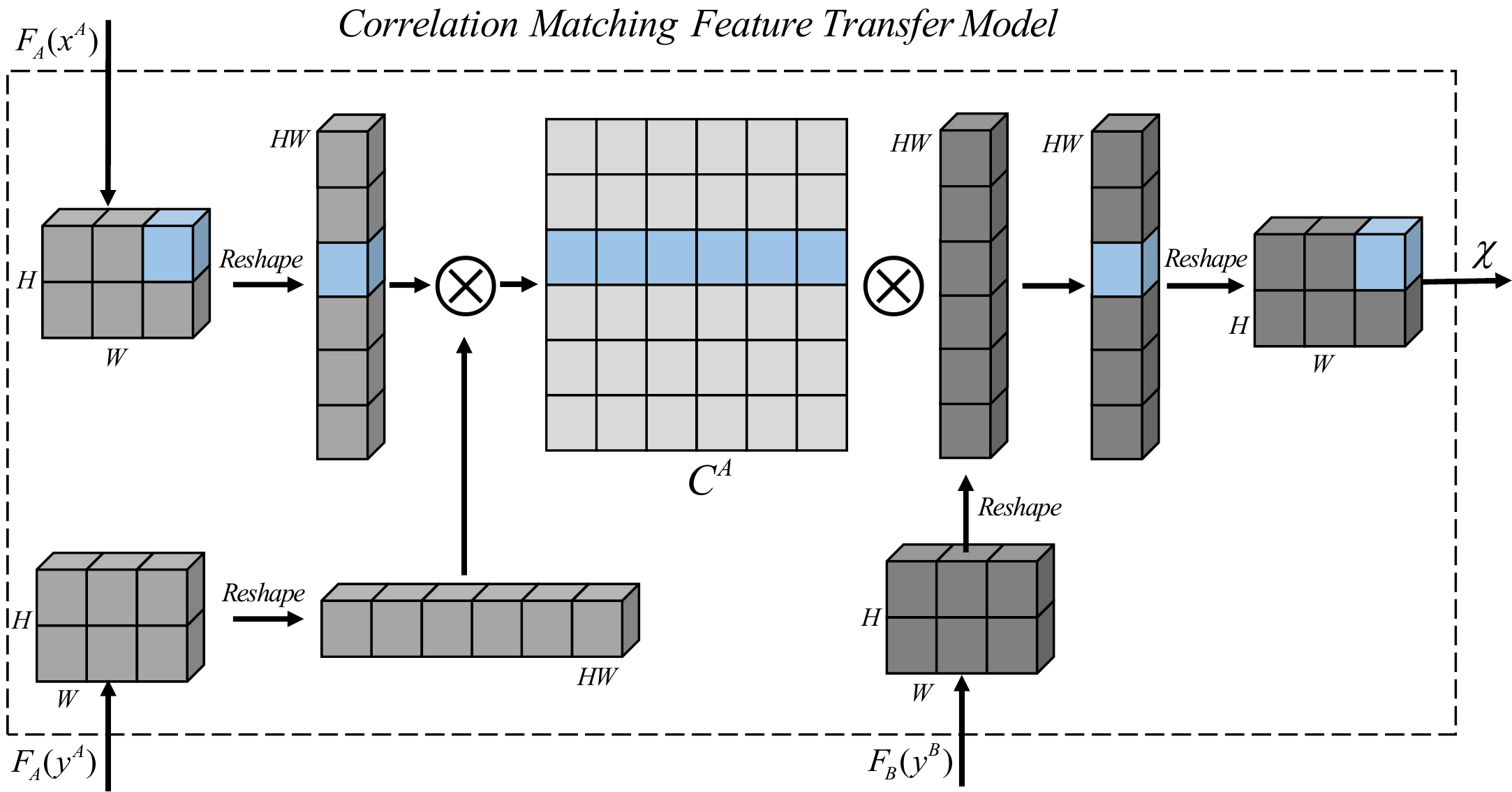}
\caption{An illustration of correlation matching feature transfer (CMFT) model.}
\label{corr}
\end{figure}
\begin{figure}[h]
\includegraphics[width=\linewidth]{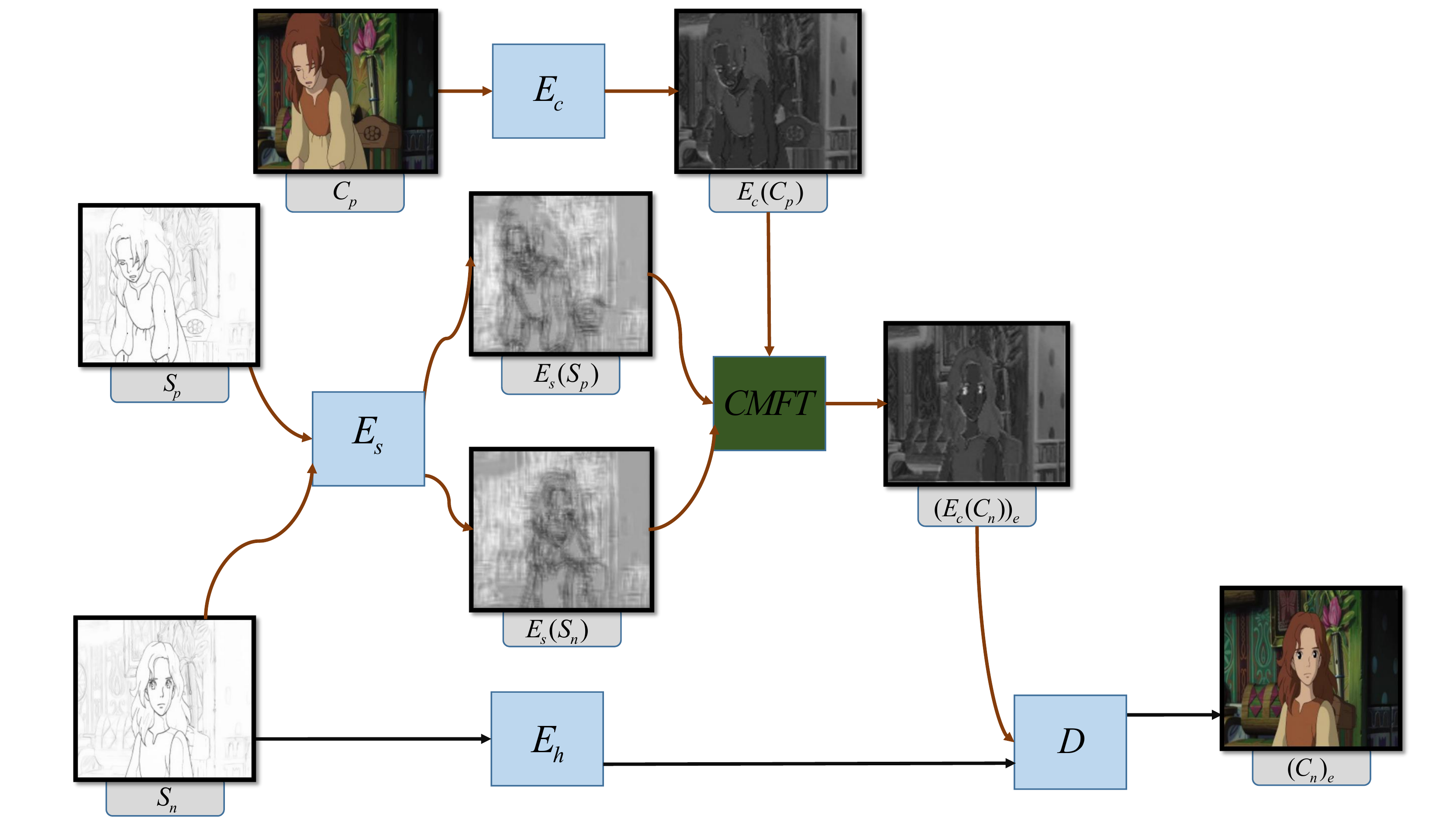}
\caption{Overview of our network. There are mainly  4 encoders and 1 decoder in our network. Encoder $E_h$ and decoder $D$ compose an U-Net structure, which is our backbone network. The previous colored image $c_p$ is encoded by $E_c$ as $E_c(C_p)$, the previous sketch and the current sketch are encoded by $E_s$ as $E_s(S_p)$ and $E_s(S_n)$ respectively. Then, in CMFT model, $E_s(S_p)$ and $E_s(S_n)$ are matched to generate a mapping matrix which is used to warp $E_c(C_p)$ to $(E_c(C_n))_e$. Taking $(E_c(C_n))_e$ as a dense estimation of $E_c(C_n)$, we reconstruct the estimation of $C_n$ by integrating $(E_c(C_n))_e$ to the decoder $D$.}
\label{overview}
\end{figure}

\begin{figure*}
\centering
\includegraphics[width=\linewidth]{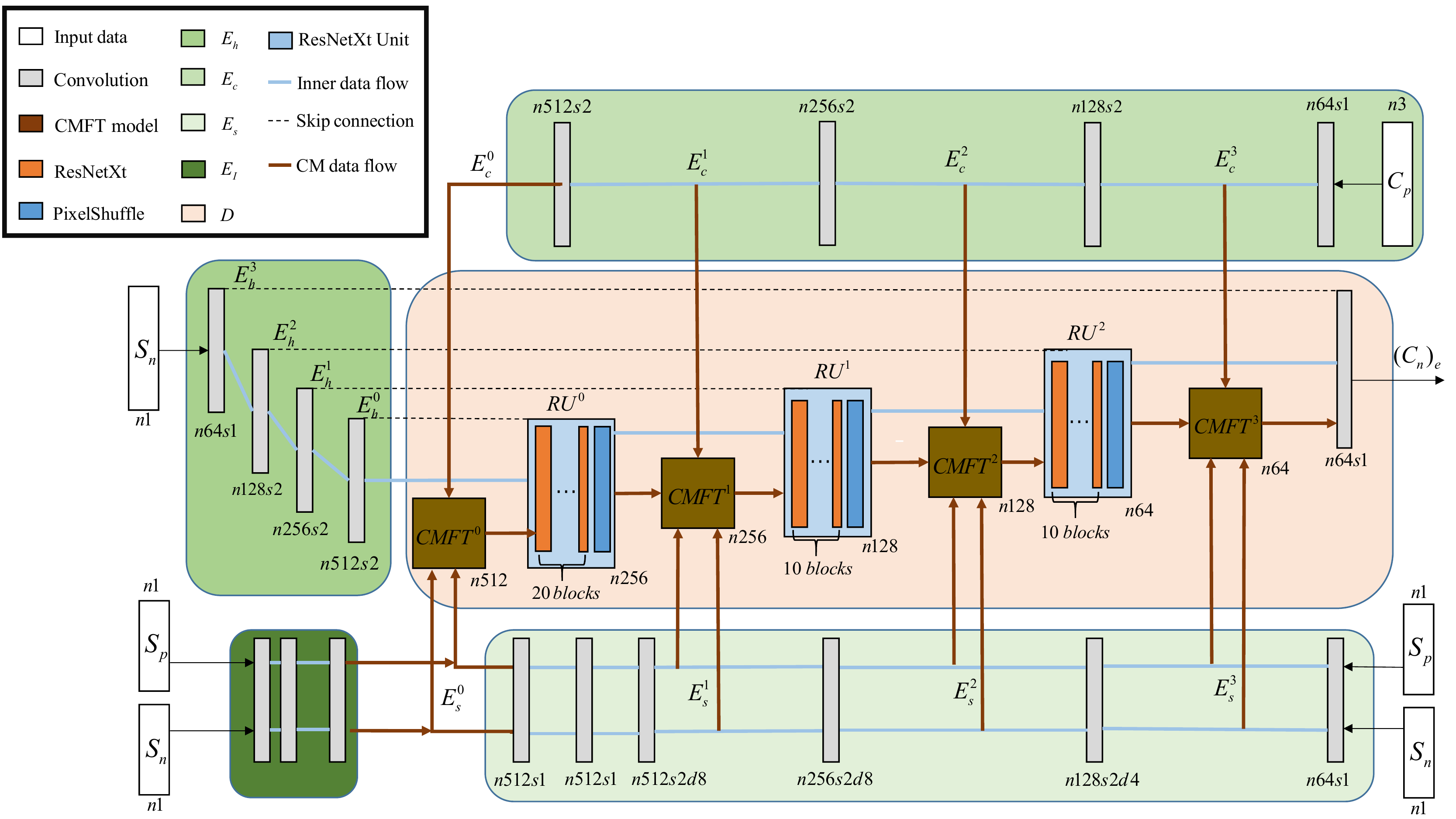}
\caption{Architecture of our generator Network with corresponding number of feature maps (n), stride (s) and dilation (d) indicated for each convolutional block. }
\label{arc}
\end{figure*}
\section{Method}
This section describes the proposed line art sequence colorization method shown in Figure \ref{arc}. 
We first build the learnable feature transfer model called correlation matching feature transfer model, which can account for consistency between frames to take into consideration temporal information.
Then, we propose the line art correlation matching feature transfer network(LCMFTN) to integrate a series of CMFT models that can act on semantic or texture features.
\subsection{Correlation Matching Feature Transfer Model}
Similar to the representation in \cite{liao2017visual}, let $x^{A}, y^{A} \in \mathbb{R}^{H^{'}\times W^{'}\times 3}$ be two images in style domain $A$, let $x^{B}, y^{B}\in \mathbb{R}^{H^{'}\times W^{'}\times 3}$ are two images in style domain $B$. We arrange $x^{A},y^{A},x^{B},y^{B}$ as image analogy $x^A:x^B::y^A:y^B$, where $x^{B}$ are unknown variable. 
This analogy implies two constraints: 1) $x^{A}$ and $x^{B}$(also $y^{A}$ and $y^{B}$) correspond at the same spatial content; 2) $x^{A}$ and $y^{A}$ (also $x^{B}$ and $y^{B}$) are similar in style (color, lighting, texture and etc). Let $F_{A}(x^{A})$, $F_{A}(y^{A})$, $F_{B}(x^{B})$, $F_{B}(y^{B})$ be the corresponding DCNN features of $x^{A}, y^{A}, x^{B}, y^{B}$, where $F_A(\cdot),F_B(\cdot) \in$ $\mathbb{R}^{H\times W\times L}$, our goal is to build a learnable network structure to find the correlation matrix $C^A \in \mathbb{R}^{HW \times HW} $ of $F_{A}(x^{A})$ and $F_{A}(y^{A})$ , then using $F_{B}(y^{B})$ and the matrix $C^A$ to transfer $F_{A}(x^{A})$ to the unknown $F_{B}(x^{B})$. 

Firstly, let $i \in HW$ and $j \in HW$ denote the index of spatial positions of the image features. Each element $C_{i,j}^{A}$ represents the correlation intensity between position $i$ in $F_A(x^A)$ and position $j$ in $F_A(y^A)$, and it is calculated  as below:
\begin{align}
\begin{split}
    C_{ij}^{A}=f(F_{A}(x^{A})_{i},F_{A}(y^{A})_{j}),\label{correlation}
\end{split}
\end{align}
in which $f$ denotes a kernel function computes the similarity of the scalars. We apply the gaussian function in this paper ($f(a,b)=e^{{a^T}b}$). As $x^{A}$ and $y^{A}$ are in the same style domain, the local pixels with the similar semantic content are similar in features, the correlation can be represented as similarities. 
Then, we estimate the feature $F_{B}(x^{B})$ by matching the pixels from $F_{B}(y^{B})$ and the estimation of $F_{B}(x^{B})$ is written as $\chi$. 
Each pixel in $\chi$ is resumed by accumulating all the pixels in $F_{B}(y^{B})$ as follows:
\begin{align}
     \chi_{i}&=\sum_{\forall j}w_{ij}F_{B}(y^{B})_{j},\label{weight}\\
	 w_{ij}&=\frac{1}{\sum_{\forall j}c_{ij}^{A}}c_{ij}^{A}, 
\end{align}
in which $w_{ij}$ denotes the weight of the pixel $j$ in $F_{B}(y^{B})$ to reconstruct the unkown feature pixel $F_{B}(x^{B})_{i}$. 
Notice that $C_{ij}^{B}$ is necessary to precisely transfer the $F_{A}(x^{A})$ to $F_{B}(x^{B})$. 
However, we replace it with $C_{ij}^{A}$ feasibly, since $x_A$ and $x_B$ (also $y_A$ and $y_B$ ) contains the same spatial content. 
Equation \ref{correlation} and \ref{weight} can be summarized as follow:
\begin{footnotesize}
\begin{align}
\begin{split}
    \chi_{i}=CMFT(F_{A}(x^{A}),F_{A}(y^{A}),F_{B}(y^{B}))_i\\=\frac{1}{\sum_{\forall j}f(F_{A}(x^{A})_i,F_{A}(y^{A})_j)}\sum_{\forall j}f(F_{A}(x^{A})_i,F_{A}(y^{A})_j)F_{B}(y^{B})_j.\label{CM}
\end{split}
\end{align}
\end{footnotesize}
Equation \ref{CM} is called correlation matching feature transfer (CMFT) model, which reconstruct the unknown $F_{B}(x^{B})$ with $F_{A}(x^{A}),F_{A}(y^{A})$ and $F_{B}(y^{B})$. 
CMFT model can be integrate to the generator of the image-to-image domain translation task.  
Different from the matching procedure in \cite{liao2017visual} and \cite{li2016combining}, the matching model will guide the learning of DCNN features.
Allowing the network to be able to learn a matching-friendly and task-friendly deep feature from the whole dataset instead of a few images will improve the robustness and accuracy for a given task. Figure \ref{corr} shows the calculation process of the CMFT model. 
In the next section, we will introduce the temporal reference in a coarse-to-fine manner by appling CMFT model to the frame colorization task.

\subsection{Line Art Correlation Matching Feature Transfer Network}
As for the coherent sketch frame colorization, the analogy can be writen as $S_{p}:C_{p}::S_{n}:C_{n}$, in which  $S_{p}$ ($S_{n}$) and $C_{p}$ ($C_{n}$) represent the
previous (current) sketch frame and previous (current) color frame respectively. 
Our generator takes $S_{n}$ as input conditioned on previous frame pair $S_{p}$ and $C_{p}$ and returns a color estimation $C_{e}$ temporally consistent to the previous colored frame.
This step can be summarized as the following formula:
\begin{align}
C_{e}=G(S_{p},C_{p},S_{n})\label{GE}.
\end{align}
U-Net\cite{ronneberger2015u} has recently been used on a variety of image-to-image translation tasks \cite{isola2017image}\cite{zhu2017unpaired}\cite{ci2018user}.
In an U-Net based network structure, the features of the encoder are directly added to the decoder by skip connections.

In our task, however, the encoding feature of the $C_{p}$ can not be directly added to the decoder for decoding $C_{n}$,
because of the spatial inconsistency. We aim to align the feature of $C_{p}$ to $C_{n}$ and add the aligned feature to the decoder in a coarse-to-fine manner with the help of the CMFT model.

As shown in Figure \ref{arc}, our generator consist of four encoders and one decoder. The backbone of our network ($E_h$ and $D$) is an U-Net based structure. 
The input of the encoder $E_{h}$ is the current sketch frame $S_{n}$, 
which contains four convolution layers that progressively halved the feature spatially from $256\times256$ to $32\times32$. 
As for the decoder $D$, inspired by  \cite{ci2018user}, we stack the ResNeXt blocks \cite{xie2017aggregated} instead of Resnet blocks \cite{he2016deep} to effectively increase the capacity of the network and use the sub-pixel convolution layers \cite{shi2016real} to increase the resolution of the features after each ResNeXt blocks. 
We represent each combination of ResNeXt blocks and sub-pixel convolution layer as $RU$ in our paper. 
Two extra encoders are introduced to encode sketches and colored images respectively, called $E_{s}$ and $E_{c}$.
 $E_{c}$ has the the same structure as $E_{h}$, and $E_{s}$ consists of 6 convolution layers. 
We add dilation \cite{yu2015multi} to some layers of $E_{s}$ to increase the receptive fields, which will enable the network to further learn some nonlocal topology features of the sketch.
Inspired by \cite{ci2018user}, we introduce a extra pre-trained sketch classification network $E_I$ to bring more abundant semantic implications to the matching process. 
We use the activations of the $6$th convolution layer of the Illustration2Vec network \cite{saito2015illustration2vec} that is pretrained on $128w$ illustrations including colored images and line art images. 
The decoder $D$ mixes the encoding features and reconstructs the color result from a coarse-to-fine manner. 
In each resolution layer of the decoder, there exists a CMFT model to accumulate the encoded features.

As shown in Figure \ref{arc}, the intermediate output of $E_{s}$ ($E_{c}$) is denoted as $E_{s}^{3},E_{s}^{2},E_{s}^{1},E_{s}^{0}$ ($E_{c}^{3},E_{c}^{2},E_{c}^{1},E_{c}^{0}$), the intermediate code of $E_{h}$ is denoted as $E_{h}^{3},E_{h}^{2},E_{h}^{1},E_{h}^{0}$.
The $CMFT$ model of each resolution is represented as $CMFT^0,CMFT^1,CMFT^2,CMFT^3$.
In the first CMFT model ($CMFT^{0}$), we aim to estimate the unknown feature $E_{c}^{0}(C_{n})$ by aligning $E_{c}^{0}(C_{p})$ in spatial domain, and we call the prediction result $(E_{c}^{0}(C_{n}))_{e}$. 
In order to make the matching more accurate and robust, we concatenate $E_{s}^{0}$ with $E_{I}$ as the matching feature so that the caculation of model $CMFT^{0}$ can be writen as Equation \ref{layer0} (we represent the concatenate operation as $ca()$).
\begin{footnotesize}
\begin{align}
\begin{split}
(E_{c}^{0}(C_{n}))_{e}\\=CMFT^{0}(ca(E_{s}^{0}(S_{n}),E_{I}(S_{n})),ca(E_{s}^{0}(S_{p}),E_{I}(S_{p})),E_{c}^{0}(c_{p}))\\=CMFT(ca(E_{s}^{0}(S_{n}),E_{I}(S_{n})),ca(E_{s}^{0}(S_{p}),E_{I}(S_{p})),E_{c}^{0}(c_{p}))
\end{split}\label{layer0}
\end{align}
\end{footnotesize}
The predicted $(E_{c}^{0}(C_{n}))_{e}$ contains the same style as $E_{c}^{0}(C_{p})$ and it is consistent with the $E_{h}^{0}(S_{n})$ in spatial domain, which makes the $(E_{c}^{0}(C_{n}))_{e}$ a good reference for the network to further construct the higher resolution features.
We concatenate the $(E_{c}^{0}(C_{n}))_{e}$ with $E_{h}^{0}(S_{n})$ and input it to the first ResnetXT upsample block ($RU^0$) to further reconstruct the higher resolution features.
We treat the output of $RU^0$ as a coarse estimation of $E_{c}^{1}(C_{n})$, so now we have the analogy as $E_{s}^{1}(S_{n})$:$RU^0$::$E_{s}^{1}(S_{p})$:$E_{c}^{1}(C_{p})$. 
We can match  $RU^0$, $E_{s}^{1}(S_{n})$ with $E_{c}^{1}(C_{p})$, $E_{s}^{1}(S_{p})$ to reconstruct a more accurate prediction $(E_{c}^{1}(C_{n}))_{e}$ and thus the calculation of $CMFT^{1}$ can be represented as Equation \ref{layer1}:
\begin{footnotesize}
\begin{align}
\begin{split}
(E_{c}^{1}(C_{n}))_{e}=CMFT^{1}(RU^0,E_{s}^{1}(S_{n}),E_{c}^{1}(C_{p}),E_{s}^{1}(S_{p}))\\=CMFT(ca(RU^0,E_{s}^{1}(S_{n})),ca(E_{c}^{1}(C_{p}),E_{s}^{1}(S_{p})),E_{c}^{1}(c_{p})). 
\end{split}\label{layer1}
\end{align}
\end{footnotesize}
Let $k$ denotes the label of the layer of the $CMFT$ model, for $k>1$. 
Since we can treat each $RU^{k-1}$ as a coarse estimation of corresponding $E_{c}^{k}(C_{n})$, the rest CMFT model ($CMFT^{2}$ and $CMFT^{3}$) can be induced from Equation of $CMFT^{1}$. 
Then, we write the calculation in an united Equation \ref{layerk}:
\begin{footnotesize}
\begin{align}
\begin{split}
(E_{c}^{k}(C_{n}))_{e}=CMFT^{k}(RU^{k-1},E_{s}^{k}(S_{n}),E_{c}^{k}(C_{p}),E_{s}^{k}(S_{p}))\\=CMFT(ca(RU^{k-1},E_{s}^{k}(S_{n})),ca(E_{c}^{k}
(C_{p}),E_{s}^{k}(S_{p})),E_{c}^{k}(c_{p}))
\end{split}\label{layerk}
,k>1.
\end{align}
\end{footnotesize}
From Equation \ref{layerk}, we can discover that the features of $C_{p}$ is aligned to $C_{n}$ in a coarse-to-fine manner. 
With the increasing of the feature resolution, more detailed information in features is considered  for matching and a more fine result can be reconstructed. At the end of the decoder, we use two convolution layers to decode the aligned features to the RGB color domain.
\subsection{Loss Objective}

\textbf{Color Loss.} We apply a color loss to the output of the generator $G$ and the
ground truth image using the following objective function:
\begin{align}
L_{1}(G)=\mathbb{E}_{X,Y}\|Y-G(X)\|_1,
\end{align}
where $X=(C_{p},S_{p},S_{n})$ and $Y=C_{n}$ \\
\textbf{Perceptual Loss.} While using only the $L1$ loss will make the generated result blurry, perceptual loss can help the model to better reconstruct fine details and edges\cite{johnson2016perceptual}. We calculate the perceptual loss on the feature maps of the VGG-19 model pre-trained on ImageNet at different depths.
\begin{align}
L_{VGG}(G)=\sum_{f \in F}\|VGG_{f}(Y)-VGG_{f}(G(X))\|_2
\end{align}
where $F$ is the set of depths of VGG-19 which are considered, in our case $F$ = 1, 3, 5, 9, 13. \\
\textbf{Objective.} By combing all the mentioned losses, the final objective function can be represented as follows:
\begin{align}
L_{G}=\lambda_{1}L_{1}+\lambda_{2}L_{VGG},
\end{align}
where $\lambda_{1},\lambda_{2}$ influence the relative importance of the different loss functions.

\subsection{Implementation Details}
The inputs of $CMFT^3$ are two size $256 \times 256$ feature maps and the shape of the relevent correlation matrix is $65536\times65536$, 
which will cause the memory overhead for a single GPU and also greatly extend the training and infering time. 
Thus we remove the $CMFT^3$ model in our implementation by directly connecting the output of $RU^2$ to the last convolution layers.

\section{experiment}
\subsection{Experimental Setup}
\textbf{Dataset.}
We collect 10 different cartoon films of Hayao Miyazaki(\textit{Howl's Moving Castle}, \textit{Whisper of the Heart}, \textit{The Wind Rises}, \textit{Ki-ki's Delivery Service}, \textit{Porco Rosso},  \textit{My Neighbor Totoro}, \textit{The Secret World of Arrietty}, \textit{Spirited Away}, \textit{Princess Mononoke}, \textit{Ponyo}),
three of which (\textit{The Secret World of Arrietty}, \textit{Whisper of the Heart}, \textit{My Neighbor Totoro}) are used for training and the rest for testing. 
We divide these training films into shots by utilizing the method described in\cite{baraldi2015shot}. 
Since frames from two different shots may not be strongly correlated and mislead the training process, we only extract training frame pairs from the same shot. 
In order to train the model to handle more diverse and intense frame variations, we design a strategy to extract more differential training pairs from a single shot. 
We apply a sliding window to every squence to obtain the frame pairs, first of which is the start frame of the window, and second of which is the last frame of the window. 
The stride of window is set to $5$, and the width is set to $40$.
In this way, we extract $60k$ pairs of training color frames and then convert this color frame set to simulate artificial line art by paintchainer's LeNet \cite{paintschainer} and take it as the sketch training set.\\
\textbf{Parameter Setting.}
Our proposed method is implemented in PyTorch, and trained and tested on a single Tesla P40 GPU.
For every experiment, we feed our network with input resized to $256\times256$ for 40 epochs, and the batch size is set to 2. 
We use the Adam optimizer with the momentum terms $b1 = 0.5$ and $b2 = 0.9$, and the initial learning rate for Adam optimizer is $1e-4$. 
For hyper-parameters setting, we fix $\lambda_{1}=10$ and $\lambda_{2}=2e-2$.\\
\textbf{Evaluation Metric.}
In order to validate results of our method, we employ Structural Similarity Index (SSIM) \cite{wang2004image} and Peak Signal to Noise Ratio (PSNR) metrics to evaluate the difference between the generated images and the ground truth frames.

\begin{table}
\begin{center}
 \scriptsize
  \caption{PSNR/SSIM result of frame sequence with stride=1}
  \label{freq1}
  \setlength{\tabcolsep}{.4mm}{
  \begin{tabular}{cccccl}
    \hline method&frame1($iv$:1)&frame2($iv$:2)&frame3($iv$:3)&frame4($iv$:4)\\
     \hline\hline
    LCMFTN&\textbf{30.24}/\textbf{0.9790}&\textbf{29.10}/\textbf{0.9747}&\textbf{28.24}/\textbf{0.9710}&\textbf{27.89}/\textbf{0.9688}\\
    LCMFTN(w/o CMFT)&29.44/0.9731&28.06/0.9675&27.28/0.9629&26.93/0.9602\\
    TCVC(our loss)&23.45/0.9086&22.78/0.9026&22.50/0.8989&22.37/0.8970\\
    TCVC&23.73/0.9164&23.05/0.9107&22.77/0.9073&22.64/0.9055\\
    Pix2Pix(with ref/our loss)&29.76/0.9593&27.98/0.9530&26.74/0.9471&26.30/0.9441\\
    Pix2Pix(with ref)&28.59/0.95594&26.82/0.9510&25.65/0.9433&25.20/0.9394\\
    DeepAnalogy&29.90/0.9773&27.22/0.9701&26.14/0.9645&25.79/0.9629\\
    \hline
\end{tabular}
}
\end{center}

\end{table}

\begin{table}
 \scriptsize
  \caption{PSNR/SSIM result of frame sequence with stride=5}
  \label{freq2}
  \setlength{\tabcolsep}{.4mm}{
  \begin{tabular}{cccccl}
    \hline method&frame1($iv$:5)&frame2($iv$:10)&frame3($iv$:15)&frame4($iv$:20)\\
    \hline\hline
    LCMFTN&\textbf{27.88}/\textbf{0.9669}&\textbf{26.84}/\textbf{0.9595}&\textbf{26.03}/\textbf{0.9539}&\textbf{25.59}/\textbf{0.9506}\\
    LCMFTN(w/o CMFT)&26.21/0.9559&25.02/0.9459&24.23/0.9388&23.73/0.9336\\
    TCVC(our loss)&21.98/0.8954&21.44/0.8872&21.04/0.8810&20.78/0.8769\\
    TCVC&22.22/0.8979&21.71/0.8905&21.30/0.8843&21.02/0.8801\\
    Pix2Pix(with ref/our loss)&25.44/0.9389&24.11/0.9274&23.25/0.9119&22.77/0.9141\\
    Pix2Pix(with ref)&24.41/0.9331&23.15/0.9196&22.35/0.9098&21.90/0.9037\\
    DeepAnalogy&24.77/0.9567&23.59/0.9462&22.67/0.9401&22.28/0.9364\\
  \hline
\end{tabular}
}
\end{table}
\begin{table}
 \scriptsize
  \caption{PSNR/SSIM result of frame sequence with stride=10}
  \label{freq3}
  \setlength{\tabcolsep}{.4mm}{
  \begin{tabular}{cccccl}
    \hline method&frame1($iv$:10)&frame2($iv$:20)&frame3($iv$:30)&frame4($iv$:40)\\
    \hline\hline
    LCMFTN&\textbf{26.84}/\textbf{0.9595}&\textbf{25.59}/\textbf{0.9506}&\textbf{24.58}/\textbf{0.9440}&\textbf{24.18}/\textbf{0.9397}\\
    LCMFTN(w/o CMFT)&25.02/0.9459&23.73/0.9336&22.24/0.9190&21.88/0.9134\\
    TCVC(our loss)&21.44/0.8872&20.78/0.8769&20.46/0.8713&20.20/0.8664\\
    TCVC&21.71/0.8905&21.02/0.8801&20.69/0.8782&20.43/0.8735\\
    Pix2Pix(with ref/our loss)&24.11/0.9274&22.77/0.9141&22.13/0.9066&21.69/0.9005\\
    Pix2Pix(with ref)&23.15/0.9196&21.90/0.9037&21.34/0.8957&20.95/0.8890\\
    DeepAnalogy&23.59/0.9462&22.28/0.9364&21.47/0.9241&21.07/0.9199\\
  \hline
\end{tabular}
}
\end{table}

\begin{table}
 \scriptsize
  \caption{Average time spent for colorize one frame}
  \label{time}
 \setlength{\tabcolsep}{1mm}{ \begin{tabular}{cccccl}
    \hline method&LCMFTN&LCMFTN(w/o CMFT)&TCVC&Pix2Pix&DeepAnalogy\\
    \hline\hline
    time(s)&0.90&0.82&0.22&0.17&7.24\\
  \hline
\end{tabular}
}
\end{table}

\begin{figure}[h]
 \label{fig1}
  \includegraphics[width=\linewidth,height=124mm]{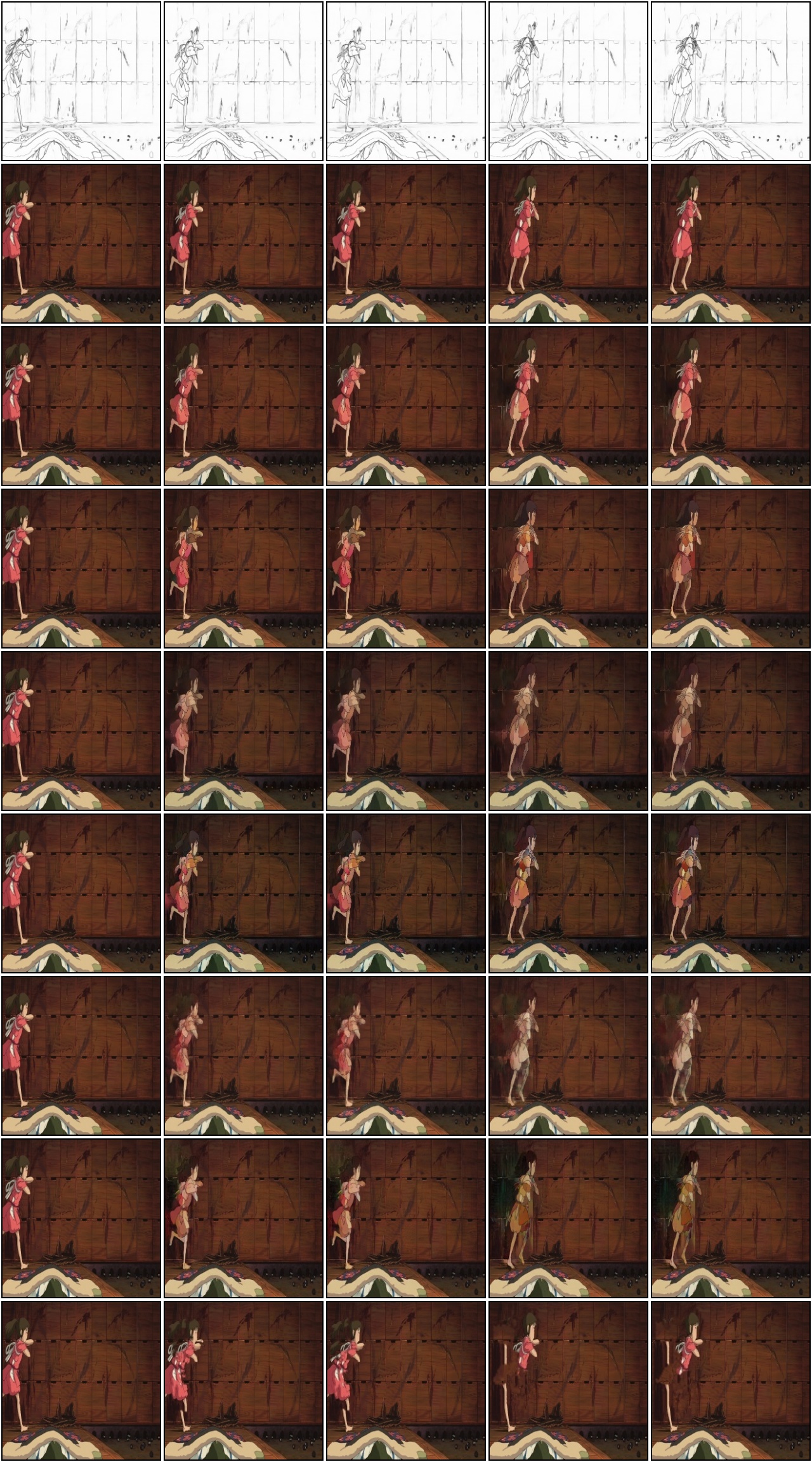}
  \caption{Example of all the compared method at stride=1, from top to bottom is the ground truth and the results of LCMFTN, LCMFTN (w/o CMFT), TCVC (our loss), TCVC, Pix2Pix (with ref/our loss), Pix2Pix (with ref), DeepAnalogy. The first colum of each rows is the origin colored image, and the successive coloum is the predicted frame conditioned on the first colored reference. The example sequence is from the film \textit{Spirited Away}.}
  \label{fig1}
\end{figure}

\begin{figure}[h]
  \includegraphics[width=\linewidth,height=124mm]{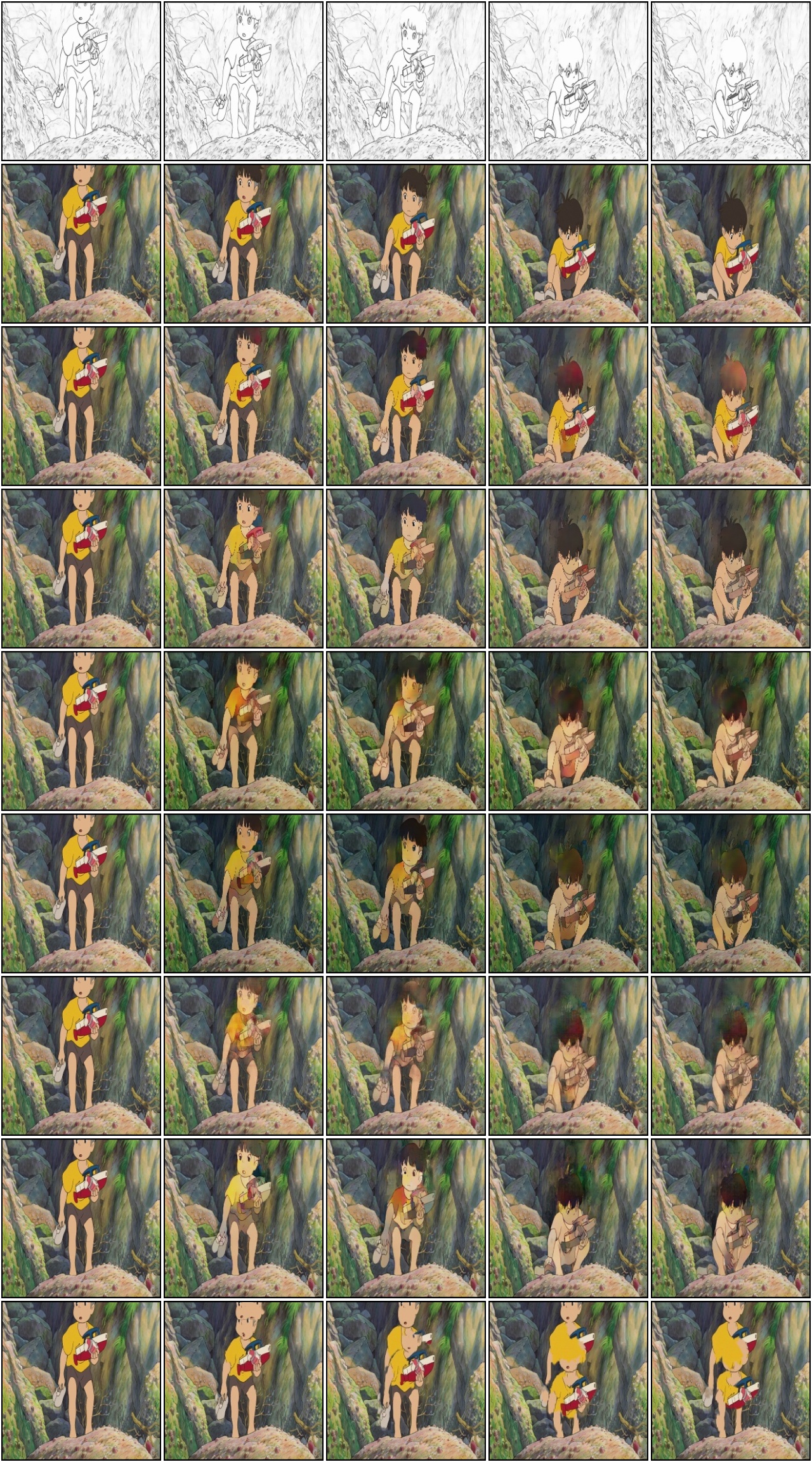} 
  \caption{Example of all the compared method at stride=5, from top to bottom is the ground truth and the results of LCMFTN, LCMFTN (w/o CMFT), TCVC (our loss), TCVC, Pix2Pix (with ref/our loss), Pix2Pix (with ref), DeepAnalogy. The first colum of each rows is the origin colored image, and the successive coloum is the predicted frame conditioned on the first colored reference. The example sequence is from the film \textit{Ponyo}. }
 \label{fig2}
\end{figure}

\begin{figure}[h]
  \includegraphics[width=\linewidth,height=124mm]{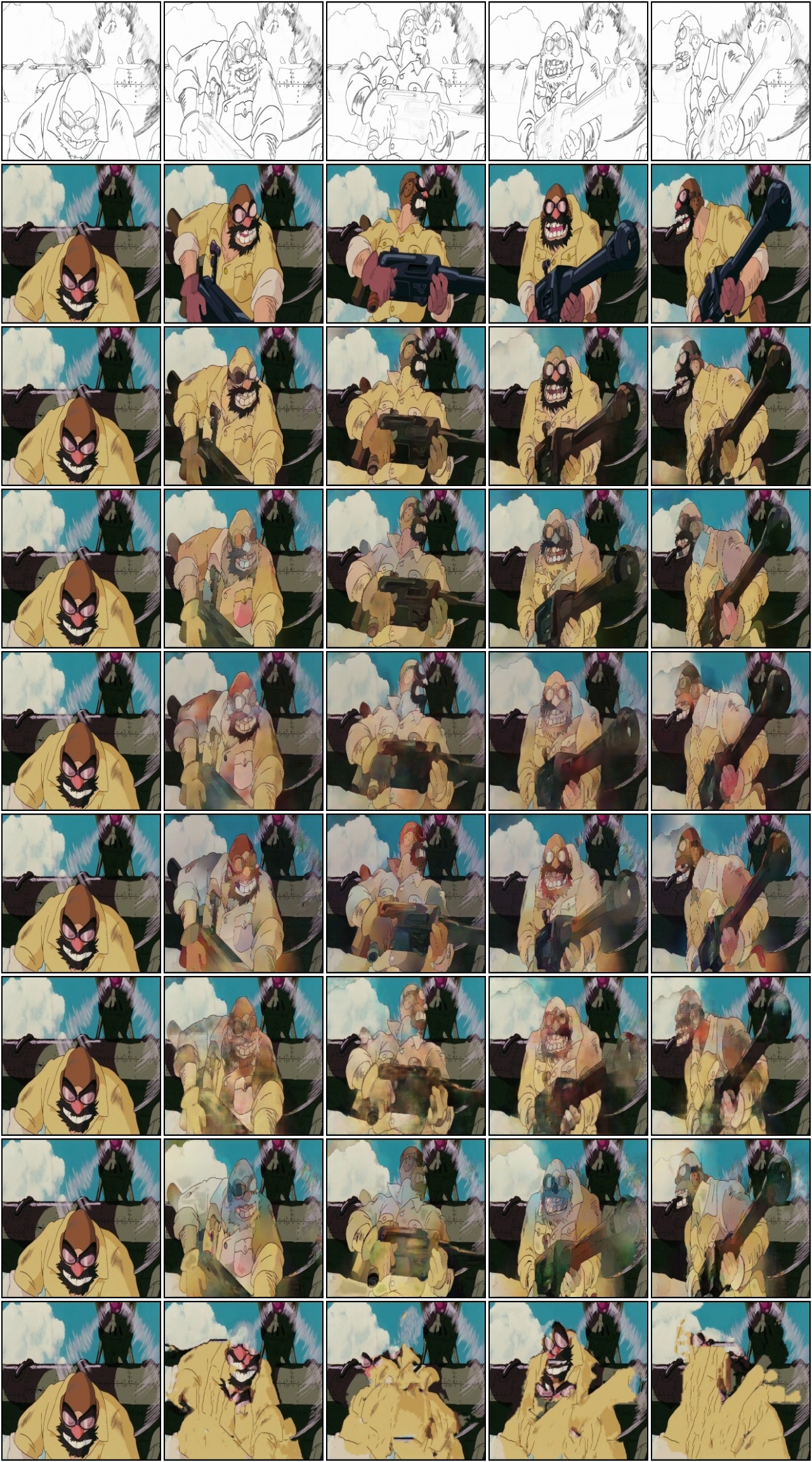}
  \caption{Example of all the compared method at stride=10, from top to bottom is the ground truth and the results of LCMFTN, LCMFTN (w/o CMFT), TCVC (our loss), TCVC, Pix2Pix (with ref/our loss), Pix2Pix (with ref), DeepAnalogy. The first colum of each rows is the origin colored image, and the successive coloum is the predicted frame conditioned on the first colored reference. The example sequence is from the film \textit{Porco Rosso}.}
  \label{fig3}
\end{figure}

\subsection{Model Analysis}
In the subsection, we investigate the influence of the CMFT model. 
We gather all shots of 7 test films into a shots set (7000 shots total). 
To see how the motion intensity and diversity influence the result, for each shot, we randomly selected 5 continuous frames at a stride of $S$, which is varied from 1 to 10. 
Obviously, the intervals between the reference frame and the generated frames range frame 1 to 40 (the interval is represented as $iv$ in tabel).
We take the first frame of this sequence as the color reference for the model to predict the successive frame. 
We eliminate the sequence from the test dataset when there exist an unchanged frame compared with the first frame, since it is not nessasery to predict the colorization when no change happens. 
We also eliminate the sequence from the test dataset when big region of uncorrected semantics shows up (for example, a character not shown in the first frame suddenly comes in in the following frames). After the clean, we get a dataset of 3500 shot sequences for testing.
Tabel \ref{freq1},\ref{freq2},\ref{freq3} have shown the result of the evaluation and Figure \ref{fig1}, \ref{fig2}, \ref{fig3} have shown the examples of the results.

To evaluate the influence of correlation mathing feature transfer model, we completely remove CMFT models from the LCMFTN network, and directly concatenate the output of each $RU$ model to the succesive $RU$. 
As shown in Table \ref{freq1}, there is a relatively smaller advantage of LCMFTN over LCMFTN (w/o CMFT) when the interval is less than 5.
This is because most of the test sketch cases only change slightly and locally between coherent frames when the interval is small, and some unknown part of frames can be easily predicted by the local ability of the network.
However, when we increase the interval to enhance the motion intensity and diversity, LCMFTN is apparently better than LCMFTN (w/o CMFT) as is shown in Table 2, 3. This is because the CMFT model is global, the correlation matrix contains similarity scores between the corresponding feature in image $x^A$ and all the features in image $y^A$ (see Figure \ref{corr}). This makes the CMFT able to learn to estimate the large transformation between coherent frames.

\subsection{Comparison against the State-of-the-Art}

We compare our method with TCVC \cite{thasarathan2019automatic}, Pix2Pix \cite{isola2017image} and DeepAnalogy \cite{liao2017visual}. 
In order to adjust the Pix2Pix model to fit example based sketch colorization task, we directly concatenate the reference to the input just as the same as the strategy introduced in TCVC.
As we can see in Table \ref{freq1}, \ref{freq2}, \ref{freq3}, the TCVC and Pix2Pix model is no better than LCMFTN both with our loss or the original loss, especially when the frame interval is big, since they are constrained by the locality of their generator. 
Since small changes between coherent sketch frames can be colorized by the local ability of U-Net, the Pix2Pix model can reach a good performance when interval=1. 
With the increasing of the stride between frames, however, the performance decreases dramatically. 
When we replace the loss of Pix2Pix to our loss, the consistency of the colorization has improved. 
This is because the GAN loss is learned from the whole data set, which will introduce some color bias when considering a single generated image. 
The results of the TCVC are unstable as some results suffer from a color inconsistency. 
As can be seen in row 5 and 6 of Figure \ref{fig1}, TCVC model tends to change the color slightly even at unchanged sketch positions.\\

The original DeepAnalogy suppose to utilize $x_A$ and $y_B$ to predict $x_B$. 
DeepAnalogy calculates the patch matching in the same image domain to guarantee the matching precision, namely, matching between DCNN features of $x_A$ and $y_A$ and DCNN features of $x_B$ and $y_B$ respectively. 
In the original version, the feature of $x_B$ ($y_A$) is estimated by fusing the feature of $x_A$ ($y_B$) and the previous layers' matching result. 
But every reference colored image has its corresponding sketch image in our task, so we eliminate the procedure of estimating the feature of $y_A$ and replace it with the real feature of $y_A$ layer-wise. 
Simultaneously, the procedure of estimating the feature of $x_B$ is still kept unchanged.  
The result of DeepAnalogy can reach a good performance when the change between frames is small (interval=1), but more matching errors show up with the increasing of motion intensity. 
Different from learnable and task-specified deep features extracted by LCMFTN, the VGG features of the sparse sketch can not provide an adequate semantic representation for the correct patch matching. 
Because of the lack of considering semantic correctness which can be learned by generator based method from abundant images in the training dataset, the result of DeepAnalogy suffers from a serious discontinuity and distortion (as can be seen in row 9 in Figure \ref{fig1}, row 9 in Figure \ref{fig3}). 
As shown in  Table \ref{time}, the calculating speed of DeepAnalogy is far slower than other methods, since the patch matching and the reconstruction of the feature of $x_B$ in each layers are both time-consuming.

\section{CONCLUSION}
In this paper, we first introduced a sketch correlation matching feature transfer model that can mine and transfer feature representations.
Then we integrated the CMFT model into a U-Net generator by designing two extra line art and colored frame encoders. 
Furthermore, we collected a  sequential colorization dataset and designed a strategy to get the training frame pair with intense and diverse variations to learn a more robust line art correlation. 
Experiments showed that our LCMFTN can effectively improve the in-between consistency and quality, expecially when big and complicated motion occurs.

{\small
\bibliographystyle{ieee_fullname}
\bibliography{art}
}

\end{document}